# Total Variation-Based Dense Depth from Multi-Camera Array

Hossein Javidnia[1], *Student, IEEE*, Peter Corcoran[2], *Fellow, IEEE*

**Abstract**
Multi-Camera arrays are increasingly employed in both consumer and industrial applications, and various passive techniques are documented to estimate depth from such camera arrays. Current depth estimation methods provide useful estimations of depth in an imaged scene but are often impractical due to significant computational requirements. This paper presents a novel framework that generates a high-quality continuous depth map from multi-camera array/light field cameras. The proposed framework utilizes analysis of the local Epipolar Plane Image (EPI) to initiate the depth estimation process. The estimated depth map is then processed using Total Variation (TV) minimization based on the Fenchel-Rockafellar duality. Evaluation of this method based on a well-known benchmark indicates that the proposed framework performs well in terms of accuracy when compared to the top-ranked depth estimation methods and a baseline algorithm. The test dataset includes both photorealistic and non-photorealistic scenes. Notably, the computational requirements required to achieve an equivalent accuracy are significantly reduced when compared to the top algorithms. As a consequence, the proposed framework is suitable for deployment in consumer and industrial applications.

## 1. Introduction

The use of consumer light field cameras such as Lytro [1], Raytrix [2] and multi-camera array in smart-phones [3-5] has received much attention in the past decade. A light field camera contains multiple viewpoints and captures the intensity of each light ray and sufficient angular information which can reveal important information about the structure of the scene.

These types of cameras have been adapted in a wide range of applications such as saliency detection [6], depth estimation [7-11], digital refocusing [12, 13] and super-resolution [14]. Recent advances in light field imaging technology enable reconstruction of scene depth in a more effective way than with conventional cameras; however, acquiring an accurate *and* dense depth map from these cameras has presented a new challenge for researchers in recent years. One of the important features of light field cameras is the ability to differentiate the rays passing through the lens which makes it easy to provide both monocular and stereo depth cues. A light field camera can extract stereo cues by capturing both magnitude and angular direction of each ray passing through the microlens while recording a scene [15]. However, in such a camera, the maximum stereo baseline is equal to the lens diameter, meaning it is often rather small.

One of the most common techniques for estimating depth from light field data is to exploit the Epipolar Plane Image (EPI) [16]. This has the advantage of being both simple to execute and fast to compute, but the accuracy is limited by the small camera baseline that is typical of these array cameras and most importantly by the illumination variation while capturing Lambertian scenes. An EPI based approach is employed in this paper to initiate the depth estimation framework. In the same way that depth estimation is performed in simple stereo image pairs, the depth from a light field set is

The research work presented here was funded under the Strategic Partnership Program of Science Foundation Ireland (SFI) and co-funded by SFI and FotoNation Ltd. Project ID: 13/SPP/I2868 on "*Next Generation Imaging for Smartphone and Embedded Platforms*".

[1] H. Javidnia is with the Department of Electronic Engineering, College of Engineering, National University of Ireland, Galway, University Road, Galway, Ireland. (e-mail: h.javidnia1@nuigalway.ie).

[2] P. Corcoran is with the Department of Electronic Engineering, College of Engineering, National University of Ireland, Galway, University Road, Galway, Ireland. (e-mail: peter.corcoran@nuigalway.ie).

computed from a set of rectified[3] images. In EPI, every pixel can be projected into a slope line which represents the depth of the corresponding scene point. The performance of applications that employ light field imaging technology is influenced by the precision of the estimated depth map. However, using only EPI to estimate depth from light field cameras introduces many challenges arising from noise in the depth map, statistical uncertainties in depth values and structural inaccuracies. We tackle these challenges by taking advantage of the Fenchel-Rockafellar duality [17] and Total Variation (TV) minimization.

Fig. 1 illustrates the schematic of a "type 1" light field camera [12] where the object is located in position (A), the camera aperture is shown in position (B) and the camera array (D) is aligned on a regular 2D grid between the main lens (C) and the image sensor (E). Each microlens located on the camera array (D), diverges the incoming light ray based on its direction. This enables the pixels underneath it to record the original rays coming from different areas of the main lens (C).

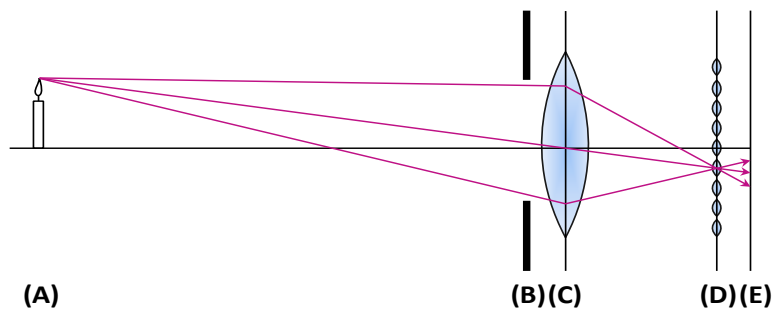

Figure 1. The schematic of a light field camera. (A): Object. (B): Camera aperture. (C): Main lens. (D): Camera array. (E): Image sensor.

Generally, a conventional pinhole camera generates an image by creating a 2D projection of a 3D scene inside a polyhedral shape as presented in Fig.2.a. The intensity of the pixel $i$ in the image plane $I$ is the intensity of the unique ray $R$ passing through the image plane and the plane containing the viewing points or the corresponding point $o$ in the object plane $O$. Whilst a pinhole camera defines a unique ray direction $R$, it is impractical because of light flux and resolution limitations. A camera with a finite lens diameter collects more light and has higher angular resolution but the intensity at any point in the image plane is now the incoherent sum of intensities from many ray directions. This drawback has been tackled by employing light field imaging techniques.

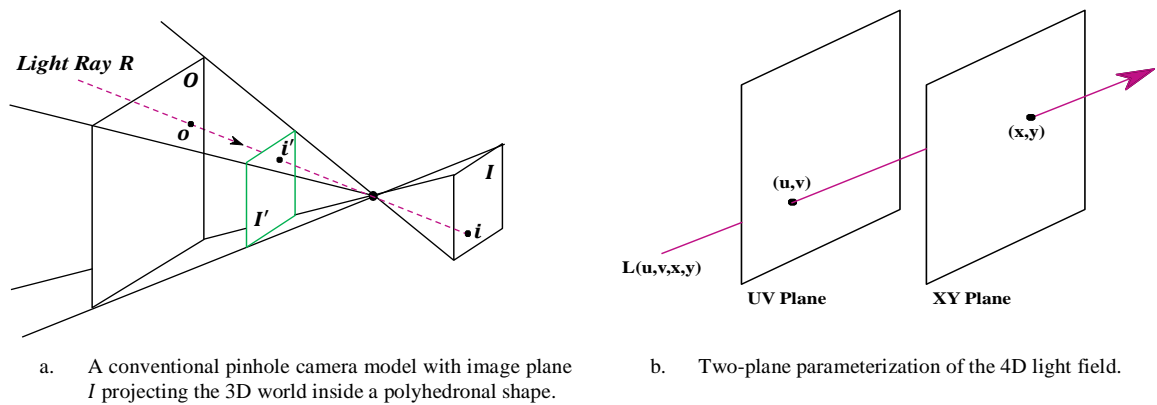

a. A conventional pinhole camera model with image plane $I$ projecting the 3D world inside a polyhedronal shape.

b. Two-plane parameterization of the 4D light field.

Figure 2. Left, the conventional pinhole camera model presented within a polyhedral shape. Right, Two-plane parameterization of the 4D light field.

---

[3] By applying rectification, all the images from the light field set are projected onto a common image plane.

Light field rendering theory explains the 4D light field data as a collection of pinhole views parallel to an image plane [18]. Commonly the position of the 2D image plane is considered as $(x, y)$ and the position of each viewing point as $(u, v)$. Fig.2.b illustrates the two-plane parameterization of the light field, proposed in [18] to simplify the plenoptic function to a 4D function. In multi-camera arrays the image sensor plane of each camera indicates $(x, y)$ and the position of the lens indicates $(u, v)$. In other words $(x, y)$ can be referred to as a pixel in the image and $(u, v)$ as the position of the camera in the array. Each pixel of the 4D light field data represents the intensity of the ray passing through image plane and the plane containing the viewing points. The light field data is stored as a 4D object as $L = (U, V, X, Y)$. Any point in the $L$ can be identified by its coordinates $[u, v, x, y]$.

In this paper, a depth estimation framework is proposed based on local EPI analysis and Total Variation (TV) minimization. The proposed minimization problem takes advantage of the Fenchel-Rockafellar duality [17]. The point in using Fenchel-Rockafellar duality [17] is that the lower bound of the minimum value will be obtained by solving the dual problem. The solution of the primal one can be found much faster by taking advantage of the information on lower bound of the minimum value.

The main contributions of this work are:

1- Introducing a lightweight computational framework to estimate depth from the 4D light field on the EPI. The proposed framework is less sensitive to occlusion, noise, spatial aliasing, angular resolution and more importantly it is 2-100 times faster/more computationally efficient than the studied state of the art methods.
2- Proposing a new computational cost function derived from the Fenchel-Rockafellar duality [17].

The rest of this paper is organized as follows. Section 2 summarizes the state of the art technology for light field depth estimation. The proposed framework is presented in detail in Section 3. The evaluation and benchmarking details are outlined in Section 4 and finally, the analyses are discussed in Section 5.

## 2. Previous Work

A lot of efforts have been made in the context of depth estimation from light field cameras including multi-view stereo matching methods [7, 19] or tensor-based methods [20, 21]. The following categories summarize different approaches on depth estimation from light field.

### 2.1. Depth from Light Field using EPI Analysis

Zhang *et al.* [22] proposed an algorithm for light field depth estimation by utilizing the linear structure of EPI [16]. The optimal slope of each line in EPI is selected from a set of candidate angles. The intensity pixel value, gradient pixel value and spatial smoothness consistency are used to aggregate the cost volume. Reliability of each pixel's disparity is identified by analyzing the matching cost curve and locally linear embedding method is used to estimate the disparity of unreliable pixels. Ma *et al*. [23] obtain the sparse depth information of the edges by exploiting local EPI analysis which is used to generate the global depth map by using regional interpolation. Yang *et al.* [24] estimate the disparity map by analyzing the EPI and detecting the slopes using the multi-label technique. Later a linear calibration method is proposed to compensate the error between the disparity values and the actual distances. Zhang *et al.* [25] proposed a spinning parallelogram operator to calculate the orientation of the EPI lines for local depth estimation. The depth estimation is based on the measurement of the slopes in EPI by maximizing the distribution distances of two parts of the parallelogram window.

Further, a confidence metric is defined to reduce the effect of the occlusions. From the approaches taken by these researchers, it is evident that the objects reflectance properties are not considered by these methods. Generally, in real scenes, the illumination is not constant over time and that introduces many challenges to depth estimation methods.

### 2.2. Occlusion-Aware Depth Estimation from Light Field

Wang *et al*. [26] proposed an occlusion aware light field depth estimation algorithm by modifying the photo-consistency condition on angular pixels. This modification along with the means or variances in the angular domain and spatial domain are used to estimate the occlusion-aware depth. In a similar approach [27] a novel data cost volume is introduced based on the correspondence and defocus cues followed by graph cut optimization to handle the occlusion in depth from the 4D light field. However, these methods struggle in handling heavy occlusions. They are mainly focused on the points which are visible in the reference view and invisible in other views.

### 2.3. Light Field Depth Estimation and Optimization

Liu *et al*. [28] tackled the light field depth estimation challenge by approaching it as an optimization problem. The objective function includes three terms as fidelity, gradient and classification. The mismatching pixels are corrected iteratively by minimizing the objective function which results in a more accurate depth map. In a similar attempt, Monteiro *et al*. [29] employed Alternating Direction Method of Multipliers to regularize the 2D EPIs and generate a dense disparity map. Unfortunately the computational time of these methods are not reported, however, their objective function contains different pixel-wise terms which introduce high computational demands and a high number of iterations to minimize.

### 2.4. Light Field Depth Estimation and Stereo Framework

Kim *et al*. [30] proposed a framework to generate stereo images from a set of light field data. Their framework is based on 3D light field and its corresponding 3D disparity volume and defines each stereo image as continues cuts through that. Graph cut optimization is also used to calculate the multi-perspective cuts. Basha *et al*. [31] used the multi-camera array for 3D reconstruction purposes by capturing a scene at two different time intervals. A 3D volume is reconstructed for each image set and the corresponding scalar volume is calculated using a nonlinear filter. The final 3D structure and motion are estimated by matching the two scalar volumes. Navarro *et al*. [32] used multi-scale and multi-window stereo method [33] to estimate disparity from two views of the light field image array. The disparity is estimated from the central view and the views in the same row and column. Later, an interpolation method is introduced based on the optical flow approach in [34] to combine the estimated disparity maps and generate the final depth map. These methods have a complex disparity constraints and it requires a high number of viewpoints. In graph-based methods, the size of the constructed graph increases significantly by adding more viewpoints to the light field set and that is a computationally intense process. On the other hand, reducing the number of viewpoints introduces notable artifacts to the depth map. In interpolation based methods, depth refinement based on optical flow formulation requires significant computational time which goes up to 1 hour and 30 minutes to process one light field image set.

### 2.5. Light Field Depth Estimation and Focal Stack Framework

Pérez *et al*. [35] proposed a focal stack frequency decomposition algorithm from light field images based on the trigonometric interpolation principle as the discrete focal stack transform. The proposed method in [35] utilizes fast discrete Fourier transform to generate refocus planes in a reasonably fast computational time. The reverse of this transformation in studied in [36] where a focal stack is used to obtain a 4D light field image set using discrete focal stack transform. Unlike [35], Mousnier *et al*. [37]

presented an approach to reconstruct 4D light field image sets from a stack of images taken by a fixed camera at different focal points. The algorithm initiates by calculating the focus map by utilizing region expansion with graph cut. Later, the depth map is estimated based on the calibration details of the camera and it is used to reconstruct the Epipolar images. The reconstructed Epipolar images are used for refocusing purposes. Lee *et al.* [38, 39] proposed a depth estimation method by separating foreground and background of the focus plane. The separated parts are converted to a binary map using the Lambertian assumption and gradient constraint. The final disparity map is estimated by accumulating the binary maps. Using the focal stack symmetry for the application of depth estimation from the 4D light field has the advantage of fast computational time, however, the final estimated depth maps suffer from sever puzzling artifact, false depth values on objects' surface, false depth values on the non-Lambertian region. Generally, the methods which are based on focal stack symmetry are highly affected by the lack of angular resolution which mostly causes false depth values on regions with a repeated pattern.

### 2.6. Physical Changes in Light Field Depth Estimation

Besides all the state of the art computational approaches for depth estimation from light field, Diebold *et al.* [40] studied the effect of light field imaging system's setup and design on the accuracy of the depth estimation. They concluded that variation in focal length and baseline of the micro cameras in an array can result in depth precision loss. It was recommended to use a precise translation stage as a good alternative for light field cameras.

## 3. Method

### 3.1. Camera Array Alignment

To generate a parallax-free array of images and eliminate the probable misalignment in the initial image sequence, we propose a generic alignment method by referring to Epipolar homography alignment. To do that, we merge all the homographies into Epipolar geometry. Considering there are $j$ plane patches in an image and their corresponding maps in the second image are characterized as:

$$\begin{aligned}\mathcal{H}_1 &= s_1 \mathcal{R}(\mathcal{I} - \mathcal{T}\mathcal{N}_1^T) \\ \mathcal{H}_2 &= s_2 \mathcal{R}(\mathcal{I} - \mathcal{T}\mathcal{N}_2^T) \\ &\dots \\ \mathcal{H}_j &= s_j \mathcal{R}(\mathcal{I} - \mathcal{T}\mathcal{N}_j^T)\end{aligned} \quad (1)$$

where $s$ is a scale factor, $\mathcal{R}$ is a $3 \times 3$ rotation matrix, $\mathcal{I}$ is the image, $\mathcal{T}$ is the second camera's translation from first camera's point of view and $\mathcal{N}(n_1, n_2, n_3)$ is the normal vector of the plane surface. So we can write:

$$\frac{s_1}{s_t}\mathcal{H}_t - \mathcal{H}_1 = s_1 \mathcal{R}\mathcal{T}\mathcal{N}_1^T - s_1 \mathcal{R}\mathcal{T}\mathcal{N}_t^T = \mathcal{K}\Delta\mathcal{N}_t^T \quad (2)$$

$$\mathcal{K} = (\kappa_1 \; \kappa_2 \; \kappa_3)^T = \mathcal{R}\mathcal{T}$$

where $\Delta\mathcal{N}_t = (\Delta n_1 \; \Delta n_2 \; \Delta n_3)^T = s_1(\mathcal{N}_1 - \mathcal{N}_t)$. Consequently it can be, concluded that:

$$d_t \mathcal{H}_t = \mathcal{H}_1 + \mathcal{K}\Delta\mathcal{N}_t^T \quad t = 2,3,\dots,j \quad (3)$$

where $d = \frac{1}{\|\mathcal{N}\|}$ is the distance of the plane from the origin and $\mathcal{H}_1$ represents the correlation between the basis homography and all the other homographies. The important feature of the Eq. (3) is that it reduces the number of independent parameters of a homography and makes them equal to the degree of freedom of a system with $j$ planar surface. Generally, a homography includes 5 *dof* indicating the camera motion and 3 *dof* representing the plane surface normal. Assuming more than one plane between two images, then $j$ homographies will have $8j$ parameters. Eq. (3) decreases the number of the parameters to $5 + 3j$ which is equivalent of the total degree of freedom in a system with $j$ planar surface.

Using Eq. (3) the motion estimation can break down into two parts:

First, considering that $\mathcal{H}_1$ and $\mathcal{K}$ are fixed, it is possible to characterize $\Delta \mathcal{N}_t$ and $\mathcal{H}_t$ by utilizing least square algorithm for each plane patches. To estimate $\Delta \mathcal{N}_t$ we define two vectors as:

$$\begin{aligned} \mathcal{U}_1 &= (\kappa_1 p_1 - \kappa_3 p_1 p_1' \quad \kappa_1 p_2 - \kappa_3 p_2 p_1' \quad \kappa_1 - \kappa_3 p_1') \\ \mathcal{U}_2 &= (\kappa_2 p_1 - \kappa_3 p_1 p_2' \quad \kappa_2 p_2 - \kappa_3 p_2 p_2' \quad \kappa_2 - \kappa_3 p_2') \end{aligned} \quad (4)$$

So $\Delta \mathcal{N}_t$ can be estimated using least squares method as:

$$\begin{aligned} \mathcal{U}_1 \Delta \mathcal{N}_t &= p_1'(h_7 p_1 + h_8 p_2 + 1) - (h_1 p_1 + h_2 p_2 + h_3) = \mathfrak{b}_{1t} \\ \mathcal{U}_2 \Delta \mathcal{N}_t &= p_2'(h_7 p_1 + h_8 p_2 + 1) - (h_4 p_1 + h_5 p_2 + h_6) = \mathfrak{b}_{2t} \end{aligned} \quad (5)$$

where $h_{1-8}$ are the parameters of the homography matrix. $(p_1, p_2, p_3)$ and $(p_1', p_2', p_3')$ are the coordinates of the point $P$ in two camera frames.

The second part is somehow the inverse process of the first part. Assuming $\Delta \mathcal{N}_t$ is fixed, $\mathcal{H}_1$ and $\mathcal{K}$ can be updated by utilizing another least square process. To estimate $\mathcal{H}_1$ and $\mathcal{K}$ we define three vectors as:

$$\begin{aligned} \mathcal{O}_t &= (p_1 \ p_2 \ 1 \ 0 \ 0 \ 0 \ -p_1 p_1' \ -p_2 p_1' \ \Delta \mathcal{N}_t P \ 0 \ -p_1' \Delta \mathcal{N}_t P) \\ \mathcal{U}_t &= (0 \ 0 \ 0 \ p_1 \ p_2 \ 1 \ -p_1 p_2' \ -p_2 p_2' \ 0 \ \Delta \mathcal{N}_t P \ -p_2' \Delta \mathcal{N}_t P) \\ \mathcal{G} &= (h_1 \ h_2 \ h_3 \ h_4 \ h_5 \ h_6 \ h_7 \ h_8 \ k_1 \ k_2 \ k_3) \end{aligned} \quad (6)$$

where $P = (p_1, p_2, p_3)$ is a point on the plane surface and $\Delta \mathcal{N}_t P = (\Delta \mathfrak{n}_{t1} p_1 + \Delta \mathfrak{n}_{t2} p_2 + \Delta \mathfrak{n}_{t3} p_3)$. So it can be concluded that $\mathcal{O}_t \mathcal{G}^T = p_1'$ and $\mathcal{U}_t \mathcal{G}^T = p_2'$. Then $\mathcal{G}$ can be estimated using least square process as:

$$\mathcal{G}^T = \frac{\mathcal{B}}{\mathcal{Q}} \quad (7)$$

where $\mathcal{B} = \begin{pmatrix} p'_{1_{11}} \\ p'_{2_{11}} \\ \vdots \\ p'_{1_{jn}} \\ p'_{2_{jn}} \end{pmatrix}$ and $\mathcal{Q} = \begin{pmatrix} \mathcal{O}_{11} \\ \mathcal{U}_{11} \\ \vdots \\ \mathcal{O}_{jn} \\ \mathcal{U}_{jn} \end{pmatrix}$. By estimating $\Delta \mathcal{N}_t$ from Eq. (5) and $\mathcal{H}_1$ and $\mathcal{K}$ using Eq. (7), one can construct the global homography from Eq. (3). The alignment process will be over when the average reprojection error is smaller than a threshold.

### 3.2. Initial Depth Estimation

Capturing a sequence of images using a multi-camera array is similar to capturing the same sequence by linearly translating one camera. Changing the camera position causes the positional changes in the image plane. Drawing out a horizontal line of constant $y^*$ in the image plane and a constant $v^*$ as the camera coordinate results in a map called EPI which can be used to visualize the positional changes in image plane. Fig.3.b shows a sample of horizontal and vertical Epipolar slices. An important feature of EPI is representing a point which is visible to all sub-aperture images by mapping it to a straight line. This feature has been used in variety of applications such as segmentation [41] and depth estimation [20]. The approach here takes advantage of this feature.

To estimate the depth map, we employed the initial part of the depth estimation algorithm of [20]; the local depth estimation on EPIs where the initial depth is constructed by using a structure tensor on each EPI $E_{y,v^*}$ ($y = 1, ..., Y$) and $E_{x,u^*}$ ($x = 1, ..., X$) to estimate the slope of the EPI lines. Two slopes are estimated for each pixel in a sub-aperture image $I_{u^*,v^*}$, one for each EPI. As illustrated in Fig.3.a the light rays $L_1$ and $L_2$ converge at point $P$ so, the following geometrical relations can be defined with regards to the depth of the point:

$$\begin{cases} x_1 + \dfrac{u_1 - x_1}{A} Z = x_2 + \dfrac{u_2 - x_2}{A} Z \Rightarrow \dfrac{\Delta u}{\Delta x} = \dfrac{Z - A}{Z} \\ y_1 + \dfrac{v_1 - y_1}{A} Z = y_2 + \dfrac{v_2 - y_2}{A} Z \Rightarrow \dfrac{\Delta v}{\Delta y} = \dfrac{Z - A}{Z} \end{cases} \quad (8)$$

where $A$ indicates the distance between two planes. $Z$ represents the depth of the point $P$ from the plane $XY$. The disparity of angular and spatial coordinates are $\Delta x = x_2 - x_1$ and $\Delta u = u_2 - u_1$, respectively. Either vertical slope $\Delta u/\Delta x$ or horizontal slope $\Delta v/\Delta y$ can be used to estimate the depth $Z$. The value of the slope is defined as the maximum pixel disparity of an object point (in pixel) among all views divided by the number of views.

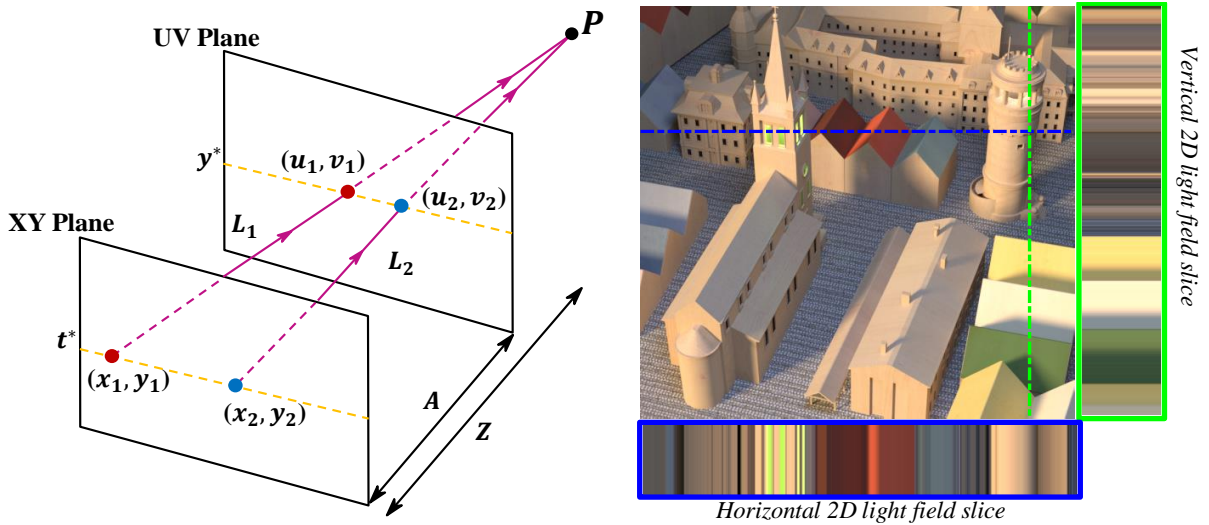

a. The depth of point $P$ can be estimated by calculating either the vertical or horizontal slope from the light field 2D slices.

b. Visualization of Epipolar image. The central view of a 9*9 camera array with horizontal and vertical 2D light field slices.

Figure 3. The concept of depth calculation from the light field using EPI analysis.

The estimated slopes are combined by minimizing the following objective function [20]:

$$F(e) = H(e) + \sum_{i=1}^{N} \int_{\Sigma_{y^*,v^*}} c_i e_i d(x, u) \quad (9)$$

where $e = (e_1, ..., e_N)$ is a vector of indicator functions, $N$ is the number of discrete depth labels, $c_i$ is the local cost function and $d$ is the depth labelling map.

Despite the fast performance of this method, it results in a noisy and unreliable depth values especially at smoother regions [42]. To tackle this issue we propose a regularization framework based on TV minimization. In the rest of this paper, the initial depth map is denoted by $D$.

### 3.3. Regularization Framework

The proposed approach can be formulated within a discrete framework. Consider a weighted graph $\mathcal{G} = (V, E)$ with vertices $v \in V$ and edges $e \in E \subseteq V \times V$, with cardinalities $n = |V|$ and $m = |E|$.

An edge passing over two vertices $v_i$ and $v_j$ is declared as $e_{i,j}$. The present paper focuses on weighted graphs which imply weights on both edges and nodes. A value assigned to each edge and node is known as edge weight $w_{i,j}$ and node weight $n_i$, respectively.

The goal is to deduce a restored vector $J$ in close proximity to the rough vector $D$ considering smooth variations of intensities inside the object.

In order to do that, the following minimization problem needs to be solved:

$$min_{x \in \mathbb{R}^n} \left( sup_{F \in B} F^T W x + \frac{1}{2\mathcal{V}}(Mx - D)^T(Mx - D) \right) \quad (10)$$

where $W$ is the weighted incidence matrix of $\mathcal{G}$ which is used to characterize the discretized gradient. $W_{ij} = 1$ if vertex $i$ is incident to edge $j$, and $W_{ij} = 0$ otherwise. $M \in \mathbb{R}^{b \times n}$ and $\mathcal{V}$ is a symmetric positive-definite weighting matrix in $\mathbb{R}^{b \times b}$. The vector $F$ is a vector representing the edges of $\mathcal{G}$. $B$ is the intersection of closed convex defined with weighted semi-norms as:

$$B = \left\{ F \in \mathbb{R}^m \mid (\forall i \in \{1, \ldots, n\}) \left\| \theta^i . F \right\|_\alpha \leq G_i \right\} \quad (11)$$

where $\| . \|_\alpha$ is the $\ell_\alpha$ norm of $\mathbb{R}^m$ with $\alpha \in [1, +\infty]$. $G = (G_i)_{1 \leq i \leq n}$ is a vector of $[0, +\infty[^n$. $\theta^i$ is a vector of multiplicative constants.

Any solution for Eq. (10) is parametrized as the optimal value corresponds to each node of the weighted graph $\mathcal{G}$.

To solve the minimization problem expressed in Eq. (10) we define the support function of $B$ as $\varrho_B$ assuming $B$ is a nonempty closed convex subset of $\mathbb{R}^n$ as:

$$\varrho_B: \mathbb{R}^n \to ]-\infty, +\infty] : a \mapsto sup_{F \in B} F^T a \quad (12)$$

This lower semi-continuous convex function is the conjugate of the indicator function $\iota_B$:

$$\iota_B = F \mapsto \begin{cases} 0, & if\ F \in B, \\ +\infty, & otherwise. \end{cases} \quad (13)$$

which leads to the modified version of the optimization function in Eq. (10):

$$min_{x \in \mathbb{R}^n} \left( \varrho_B(Wx) + \frac{1}{2\mathcal{V}}(Mx - D)^T(Mx - D) + \frac{\daleth \|Zx\|^2}{2} \right) \quad (14)$$

where $\daleth \in ]0, +\infty[$ and $Z \in \mathbb{R}^{n \times n}$ is the projection matrix onto the nullspace of the $M$.

Eq. (14) can become equivalent to Eq. (10) where $M$ is injective. The term $x \mapsto \frac{\daleth \|Zx\|^2}{2}$ vanishes when $M$ is injective. However, it helps the objective function to stay convex by bringing an additional regularization term when $M$ is not injective. Assuming $B$ is a nonempty closed convex then Eq. (14) acknowledges a distinctive solution. In this case, Eq. (14) can be redefined based on Fenchel-Rockafellar duality theorem [17] as:

$$min_F \phi(F) + \iota_B(F) \qquad (15)$$

where $\phi: F \mapsto \frac{F^T W \gamma W^T F}{2} - \frac{F^T W \gamma M^T D}{\mathcal{V}}$ and $\gamma = \frac{\mathcal{V}}{M^T M + \daleth Z}$. The optimum solution $\hat{x}$ of Eq. (14) is concluded from each optimum solution $\hat{F}$ of the duality problem in Eq. (15) by the following relation:

$$\hat{x} = \gamma(\frac{M^T D}{\mathcal{V}} - W^T \hat{F})) \qquad (16)$$

The indicator function $\iota_B$ can be broken down into the sum of indicator functions of the convex subsets. Consequently, the Fenchel-Rockafellar duality [17] in Eq. (15) can be re-written as:

$$min_{F \in \mathbb{R}^m} \sum_{q=1}^{e} \iota_{B_q}(F) + \phi(F) \qquad (17)$$

where for each set $B_q$, $q \in \{1, ..., e\}$. The above convex function is optimized by employing Parallel Proximal algorithm [43] as shown in Algorithm 1.

---

**Algorithm 1:** General form of Parallel proximal algorithm
---
Set $\lambda \in [0, +\infty], \lambda_\ell \in [0,2]$
For $q \in \{1, ..., e\}$ set $(w_q)_{1 \leq q \leq e} \in [0,1]$
Set $(y_{q,0})_{1 \leq q \leq e} \in (\mathbb{R}^m)^e$
For $i = 0: ...$
  For $q = 1: e$
    $\wp_{q,i} = prox_{\lambda D_q / w_q} y_{q,i} + \alpha_{q,i}$
  $\wp_i = \sum_{q=1}^{e} w_q \wp_{q,i}$
  For $q = 1: e$
    $y_{q,i+1} = y_{q,i} + \lambda_\ell(\wp_i - F_i - \wp_{q,i})$
  $F_{i+1} = F_i + \frac{\lambda_\ell(\wp_i - F_i)}{2}$

$\lambda$ and $w_q$ are the positive regularization parameter and a positive constant, respectively. Beside the possible error term $\alpha$, a relaxation parameter $\lambda_\ell$ is defined in each iteration.

### 4. Evaluation

The evaluation of the technique proposed in this paper is performed in two parts and is based on HCI, the Heidelberg 4D Light Field Benchmark [44]. The first part presents the evaluation of the optimization function and the second part compares the accuracy of the estimated depth map against the state of the art methods which are ranked in HCI, Heidelberg benchmark. The evaluation is performed using the standard "evaluation package/toolkit" provided by the benchmark to assess the performance of the proposed framework.

This benchmark is the first in the state of the art which provides light field image sets with ground truth data and standardized the evaluation framework. The light field image sets are designed to challenge accuracy and reliability of different algorithms in different aspects such as occlusion handling, performance on convex versus concave geometry, keeping fine structure and etc.

The current version of the benchmark provides $9 \times 9 \times 512 \times 512 \times 3$ light field images along with corresponding camera configuration files. The benchmark contains 3 sets including Stratified, Test and Training. These categories are pre-defined in the benchmark. The Stratified set contains 4 light field image sets as shown in Fig.4.a-Fig.4.d. The goal of the Stratified set is to introduce challenges which can lead to fine-tuning algorithm parameters and performance metrics for real-world images. The Training set contains 4 light field photorealistic image sets as illustrated in Fig.4.e-Fig.4.h. The goal of this set is to evaluate the performance of algorithms on respecting scene structures, handling complex occlusions, slanted planar surfaces, and continuous non-planar surfaces.

In this paper, Stratified and Training sets are used for comparison purposes which include 8 different light field image sets with different configurations. The ground truth data for all these image sets is provided by the HCI benchmark.

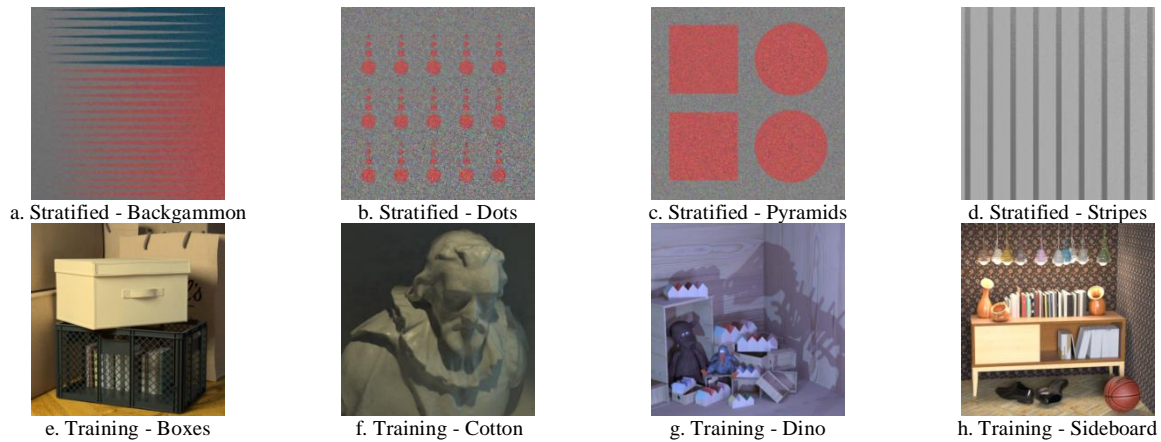

a. Stratified - Backgammon    b. Stratified - Dots    c. Stratified - Pyramids    d. Stratified - Stripes

e. Training - Boxes    f. Training - Cotton    g. Training - Dino    h. Training - Sideboard

Figure 4. 4D Light field image sets used for evaluation. First row shows the four stratified scenes and the second row shows the four photorealistic training scenes.

### 4.1. Residual Norm Evaluation

This section presents the convergence analysis of the optimization function. The maximum number of iterations $i$ and the regularization parameter $\lambda$ in Algorithm 1 are set to 300 and 0.5 for light field sets used in this paper. These values are chosen experimentally and for the evaluated image sets, they provide the average best results. Fig.5 illustrates the residual norms of the optimization function for each light field image set in Stratified and Training sets. As shown in this figure, the presented optimization method, results in a considerably low convergence error after ~50 iterations. The average convergence error of 0.01 at iteration 50 outlines the fast performance of the optimization function. Residual norms are used to verify a solution to the optimization function by substituting it into the function. The residual vanishes when the optimal solution is found [45, 46].

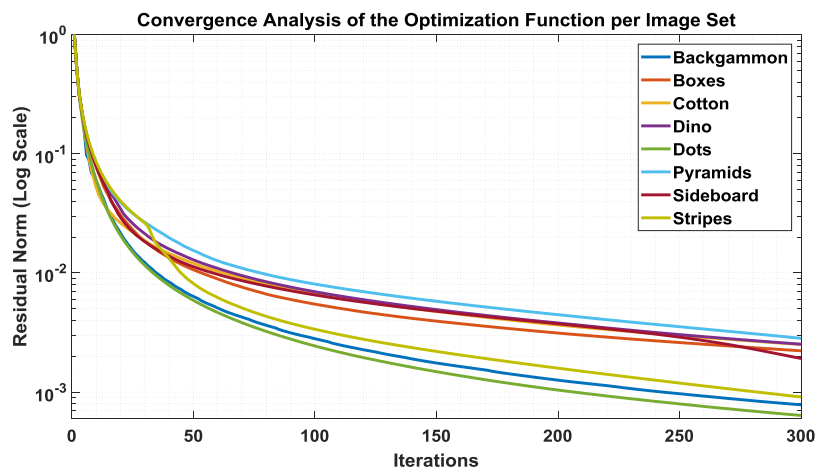

Figure 5. Residual norm analyses of the minimization function for stratified and training sets. The maximum iteration number is set to 300 for all scenes.

### 4.2. Disparity Estimation Evaluation

In this section, the accuracy of the estimated disparity maps are compared against the provided ground truth, 6 top algorithms ranked by the benchmark and one baseline algorithm. The best estimated disparity maps from all these algorithms are provided by their authors and the benchmark. The top algorithms are chosen based on the average value of the BadPix(0.03) metric and include OBER-cross+ANP [47], SPO-MO [47], OBER-cross [47], OFSY_330/DNR [48], PS_RF [47] and SPO [25]. The baseline algorithm is EPI2 [20] (the local depth estimation on EPIs) which is used to provide the initial depth map in this paper. As this research is more focused on generating accurate depth map and increasing its accuracy, three metrics including BadPix(0.07), MSE and Q25 are chosen for comparison purposes. These metrics are categorized as "High accuracy metrics" [49] in this benchmark.

The BadPix(0.07) is quantified as:

$$BadPix_M(0.07) = \frac{|\{x \in M : |d(x) - gt(x)| > 0.07\}|}{|M|} \quad (11)$$

where $d$ is the estimated disparity map, $gt$ is the ground truth disparity map and $M$ is the evaluation mask. BadPix(0.07) shows the percentage of pixels at the given mask with $|d - gt| > 0.07$. The error threshold "0.07" is the default value defined by the benchmark.

MSE shows the mean squared error over all pixels at the given mask, multiplied with 100:

$$MSE_M = \frac{\sum_{x \in M}(d(x) - gt(x))^2}{|M|} \times 100 \quad (12)$$

Q25 represents the maximum absolute disparity error of the best 25% of pixels for each algorithm, multiplied by 100.

Fig. 6 and Fig. 7 visualize the distribution of the accurate pixels and mean square error of the proposed method and the top state of the art algorithms for Stratified and Training sets, respectively. Each column in these figures shows the result of an algorithm and each row visualizes a metric. For all the metrics the lower values show a better result. These figures illustrate the performance of the proposed method based on the high accuracy metrics while dealing with different noise level, complex occlusions, slanted planar surfaces and complex scene structure.

In BadPix(0.07) metric, the good pixels are shown in green and the faulty ones are presented in red. In MSE, the correct pixel values are shown in white, the pixels with too close values are illustrated in blue and the pixels with too far values are shown in red. In Q25, the white/yellow parts indicate the good and the red parts indicate relatively bad pixels.

The fluctuation in the ranking of the algorithms based on each metric raise from their differences in terms of data and final optimization term. Some of these algorithms such as EPI2, SPO and OBER-cross estimate the disparity based on EPI analysis. OFSY_330/DNR utilizes the focal stack symmetry for disparity estimation and PS_RF uses the multi-view stereo approach by building individual cost volumes for its data terms.

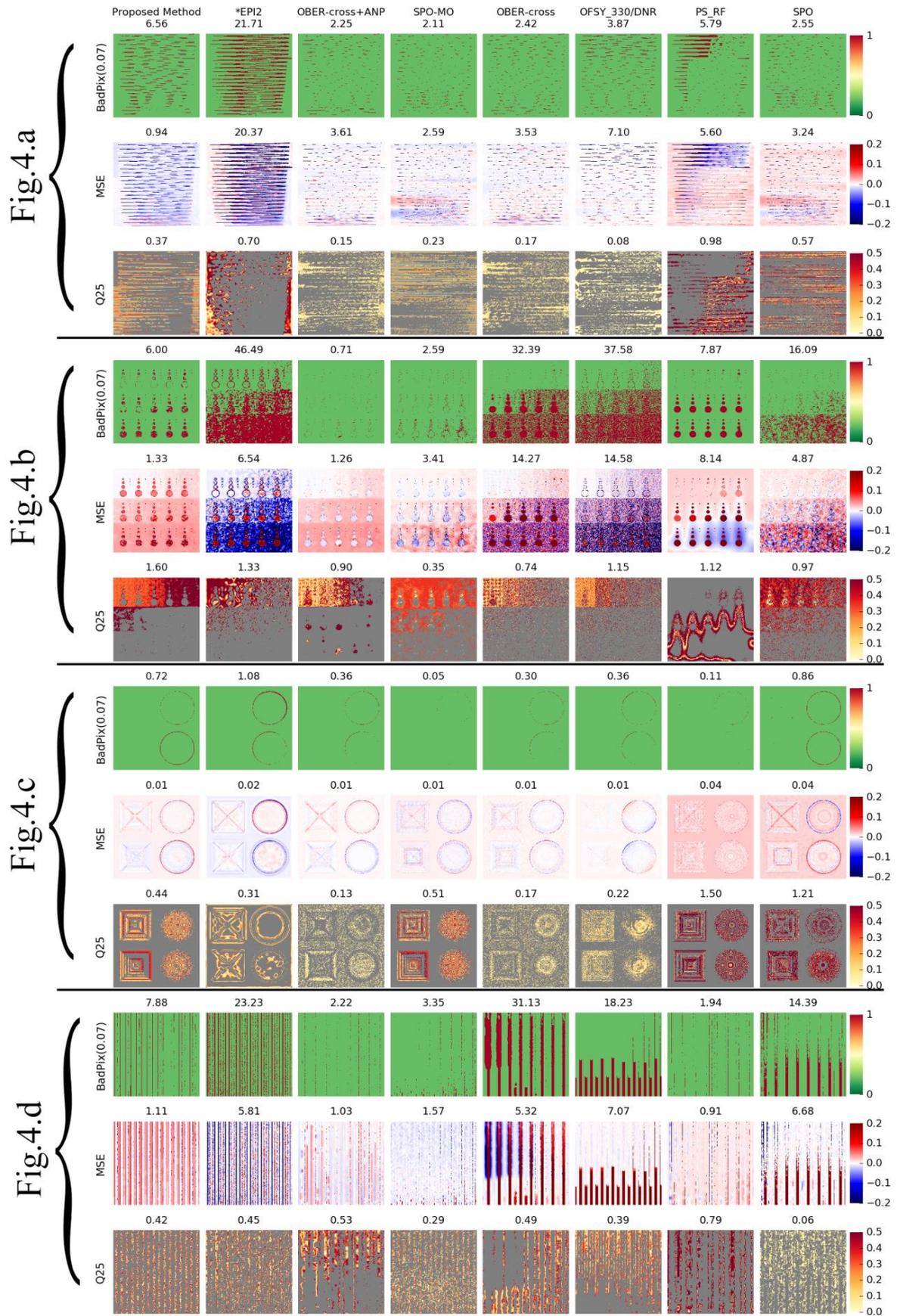

Figure 6. Stratified Scenes-Visualizations of BadPix(0.07), MSE and Q25 error metrics per algorithm are shown for the proposed method, a baseline and six most accurate algorithms on the Stratified set. Each column represents an algorithm. The rows with BadPix(0.07) show the percentage of pixels at the given mask with $|d - gt| > 0.07$. The row with MSE label show the mean square error map and the row with Q25 label shows the absolute error of the 25% of the best pixels for each algorithm

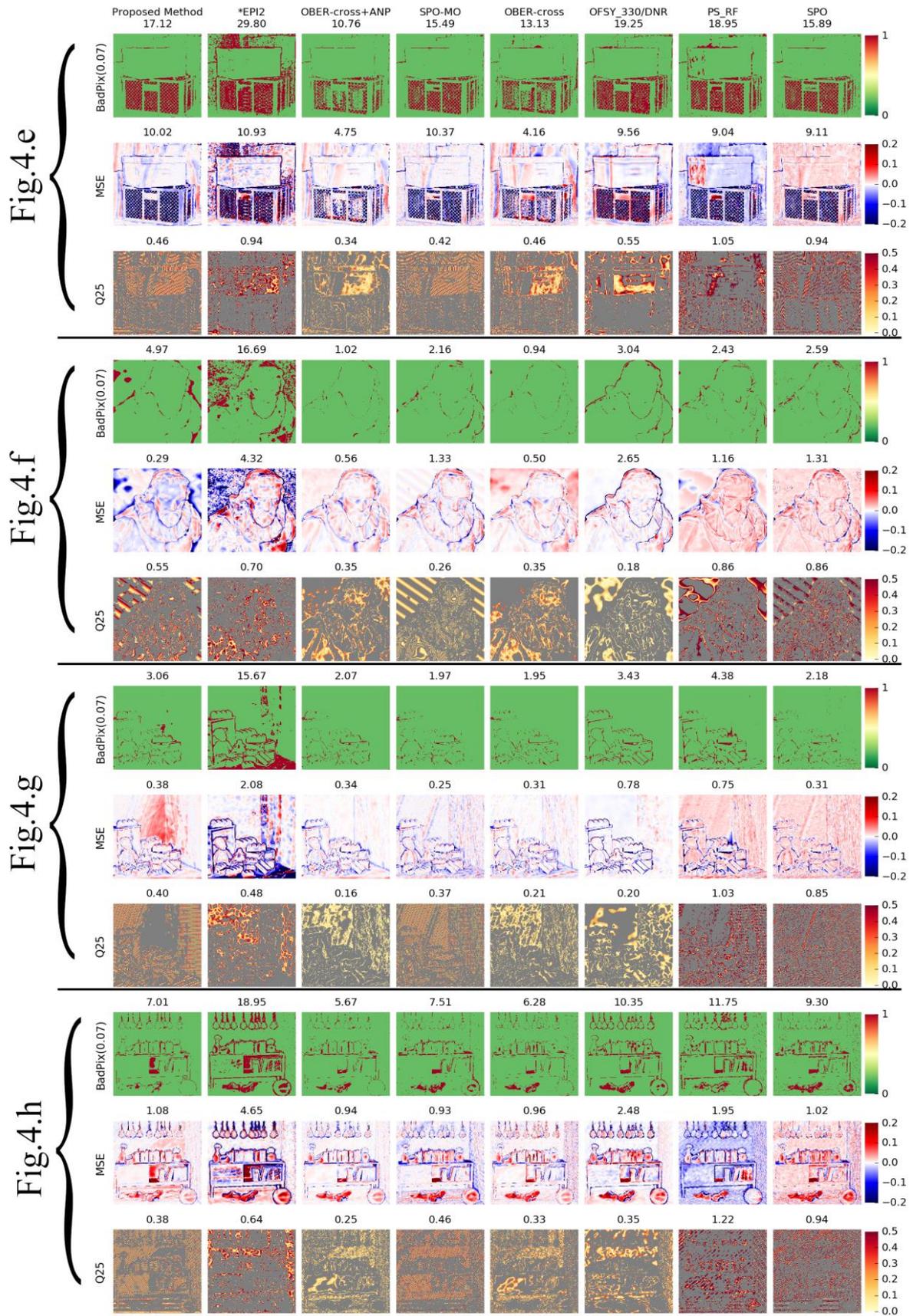

Figure 7. Training Scenes-Visualizations of BadPix(0.07), MSE and Q25 error metrics per algorithm are shown for the proposed method, a baseline and six most accurate algorithms on the Training set. Each column represents an algorithm. The rows with BadPix(0.07) show the percentage of pixels at the given mask with $|d - gt| > 0.07$. The row with MSE label show the mean square error map and the row with Q25 label shows the absolute error of the 25% of the best pixels for each algorithm

According to the evaluations and as shown in Fig. 6 and Fig. 7 there is no single best algorithm to be considered as the superior one. As it is challenging from Fig. 6 and Fig. 7 to understand the relative merits of each technique we provide Table 1 and Table 2 which outline the average numerical values per metric per algorithm for the images in each set in both stratified and photorealistic scenes, respectively. The values presented in these tables outline the close performance of the proposed method in comparison to the top state of the art algorithms. The cells are color encoded in each row based on the ranking of each method per metric. The green represents the best performing method and red shows the poorest performing one. These values indicate that the method proposed in this work can estimate depth maps with accuracy very close to the most accurate methods in the benchmark. The proposed method provides a reduction of ~56.5% averaged across all the error metrics when compared to the baseline algorithm EPI2.

Table 1. Average values of metric per algorithm for the images in Stratified set

|  | Proposed Method | EPI2 | OBER-cross+ANP | SPO-MO | OBER-cross | OFSY_330/DNR | PS_RF | SPO |
|---|---|---|---|---|---|---|---|---|
| BadPix(0.07) | 5.29 | 23.12 | 1.38 | 2.02 | 16.56 | 15.01 | 3.92 | 8.47 |
| MSE | 0.84 | 8.18 | 1.47 | 1.89 | 5.78 | 7.19 | 3.67 | 3.7 |
| Q25 | 0.707 | 0.69 | 0.42 | 0.31 | 0.39 | 0.46 | 1.09 | 0.702 |

Table 2. Average values of metric per algorithm for the images in Training set

|  | Proposed Method | EPI2 | OBER-cross+ANP | SPO-MO | OBER-cross | OFSY_330/DNR | PS_RF | SPO |
|---|---|---|---|---|---|---|---|---|
| BadPix(0.07) | 8.04 | 20.27 | 4.88 | 6.78 | 5.57 | 9.017 | 9.37 | 7.49 |
| MSE | 2.94 | 5.49 | 1.64 | 3.22 | 1.48 | 3.86 | 3.22 | 2.93 |
| Q25 | 0.44 | 0.69 | 0.27 | 0.37 | 0.33 | 0.32 | 1.04 | 0.89 |

Fig. 8 represents the difficulty of each scene type as a heatmap. Each pixel in the heatmap represents the percentage of pixels with the disparity error > 0.07 pixels averaged across all of the algorithms. Thus more than 90% of these algorithms struggle in detecting the correct disparity value for the pixels inside the box in the "Boxes" image set as shown in the first image in Fig. 8.

Another example is the "Sideboard" image set where 80-95% of the algorithms struggle with estimating the correct disparity on the surface of the shoes. The brighter parts in this figure indicate challenging areas. The "Backgammon" scene challenges the algorithms in occlusions and keeping fine structure and the "Stripes" set evaluates the methods for handling textured occlusion boundaries.

Using "Dots" image set, the robustness of each algorithm is evaluated against camera noise. The heatmap for the "Dots" image set shows that almost all the algorithms are sensitive to noise. The bottom row of this image set indicates that about 40-50% of the algorithms have a problem in detecting the correct disparity for the background objects while 70-80% of the algorithms struggle in detecting correct disparity values for the foreground objects in presence of noise. The performance of these algorithms is challenged in terms of processing convex, concave, rounded and planar geometry in the "Pyramids" image set.

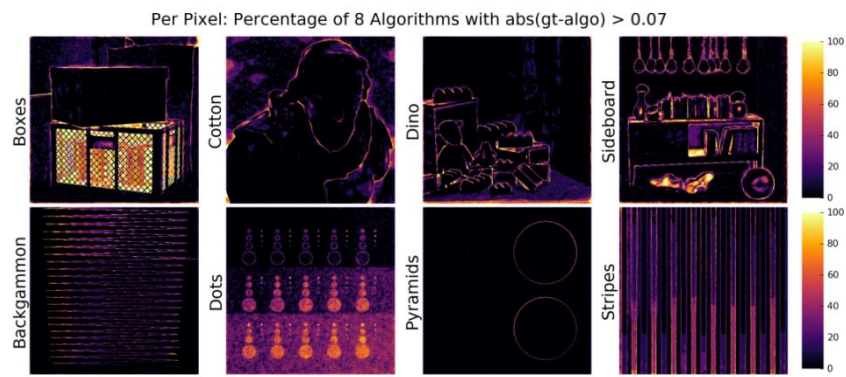

Figure 8. Scene difficulties visualized as heatmaps.

Fig. 9 presents the percentage of pixels with correct disparity on the Stratified and Training scenes for the increasing error thresholds. The PerPixBest [49] in Fig. 9 is an artificial algorithm made for evaluation purposes. "The lowest absolute disparity difference to the ground truth at each individual pixel among all algorithms" [49] is used to create the PerPixBest metric.

The algorithms ranking change by varying the thresholds for absolute disparity error. The difference in performance of the algorithms for high error thresholds is relatively small. The lower thresholds show a more apparent difference in the performances and the ranking of the algorithms change significantly for the thresholds between 0.010 and 0.032. Despite the weak tolerance in the performance of the proposed method in lower thresholds between 0 and 0.035, it is ranked among the top three methods from the threshold 0.048 onwards. OFSY_330/DNR has the best performance in lower error thresholds between 0 and 0.012 and its performance reduces for the thresholds higher than 0.012. OBER-cross+ANP has the second best performance up to the threshold 0.022 and it achieves the best performance from the error threshold 0.034 onward while competing very closely with SPO-MO.

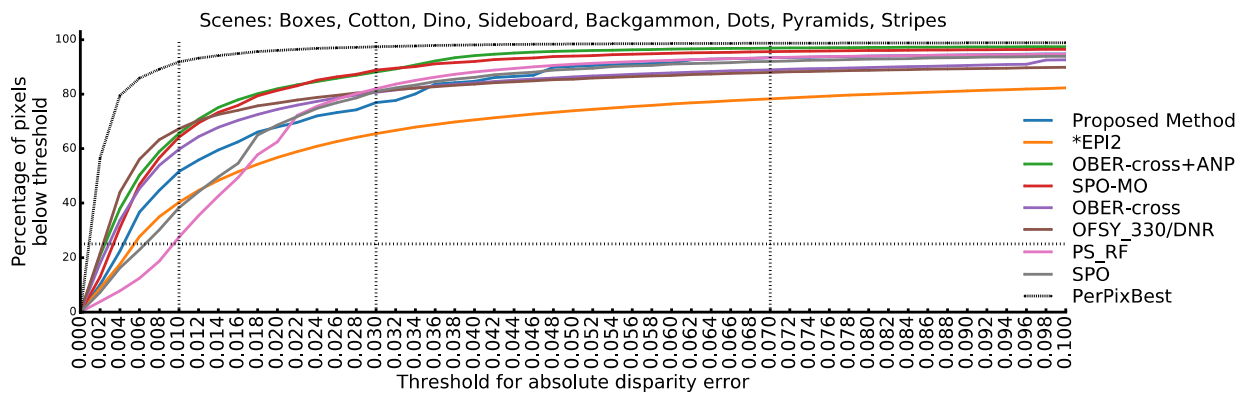

Figure 9. The percentage of pixels with correct disparity on the photorealistic and non- photorealistic scenes with increasing error thresholds.

Fig. 10 and Fig. 11 illustrate the disparity maps, ground truth error map and the median error map of the studied algorithms for Stratified and Training sets, respectively. Each row in these figures represents an algorithm. For each algorithm per individual image set, there are three columns illustrating the disparity maps, ground truth error map and the median error map. To generate the median error map, the median of the absolute disparity differences of all algorithms with the ground truth is computed for each pixel. Further, the absolute disparity difference of each algorithm is subtracted from the median error. The median error map gives a conceptual understanding of the parts of the image where algorithms perform below or above average performance of all algorithms. Yellow parts in this map represent the average, green above-average and red below-average performance.

The median error maps of the proposed method in Fig. 10 indicate its close performance to the average performance of all algorithms in well-structured scenes, complex occlusions and different noise levels. The same maps in Fig. 11 show how competitive the proposed method performs compared to the other algorithms while dealing with slanted planar surfaces and complex scene structure. However, there are still highly textured areas with fine patterns such as box frames in the "Boxes" image set which introduces many challenges to depth estimation algorithms. For instance, OBER-cross+ANP and OBER-cross algorithms estimated the disparity of the pixels beyond the box frames in "Boxes" image set above-average of the median algorithm. On the other side, EPI2 estimated the disparity level for the same area highly below-average of the median algorithm.

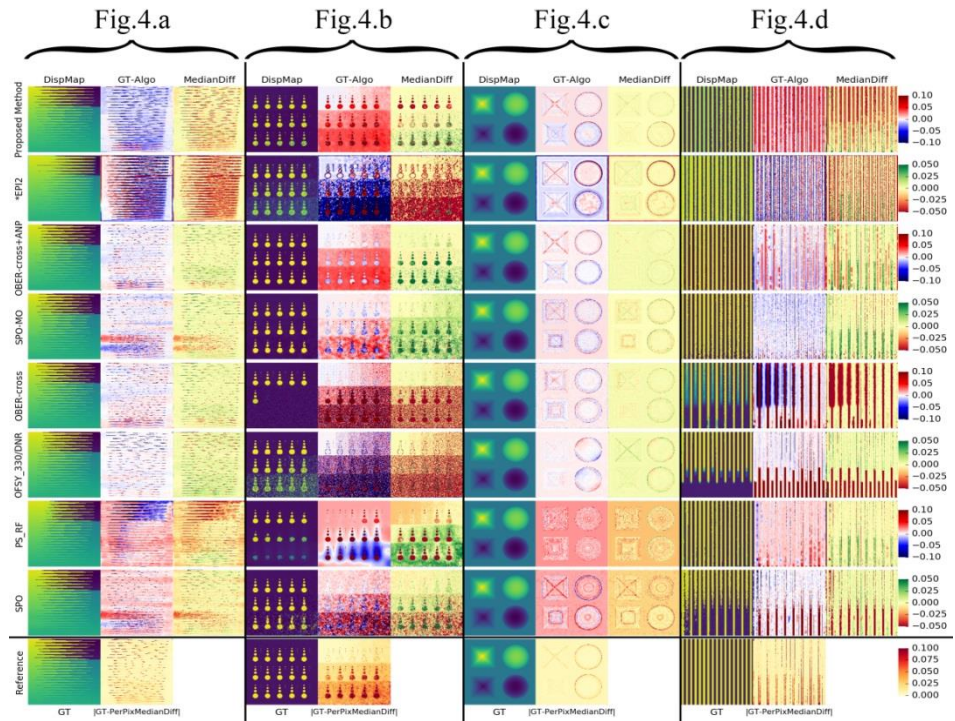

Figure 10. Stratified Scenes-Visualizations for disparity maps and their differences with ground truth. Each row represents an algorithm. The first column for each stratified scenes illustrates the disparity maps of the proposed method and the studied algorithms. The second column illustrates the disparity difference to the ground truth. Highly accurate parts are shown in white, too close in blue and too far in red areas. The third column illustrates how algorithms perform relative to the median algorithm performance. Yellow parts show average, green above-average and red below-average performance. The last row of the figure illustrates the ground truth disparity maps and the median absolute disparity difference to the ground truth at each individual pixel among all algorithms.

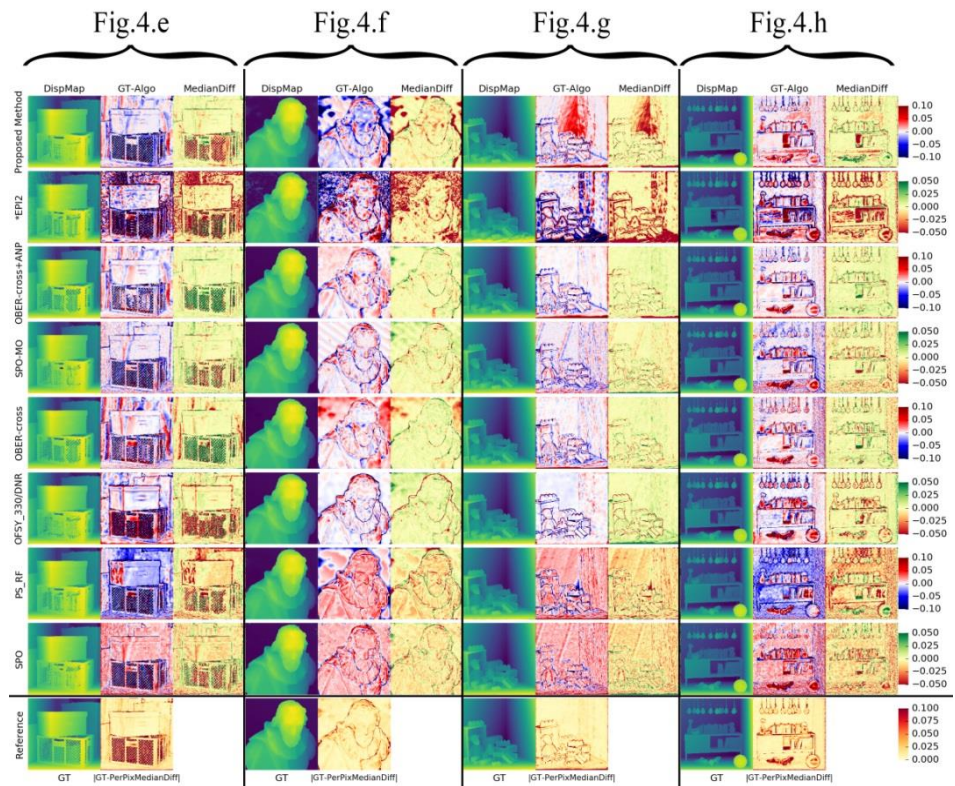

Figure 11. Training scenes-Visualizations for disparity maps and their differences with ground truth. Each row represents an algorithm. The first column for each training scenes illustrates the disparity maps of the proposed method and the studied algorithms. The second column illustrates the disparity difference to the ground truth. Highly accurate parts are shown in white, too close in blue and too far in red areas. The third column illustrates how algorithms perform relative to the median algorithm performance. Yellow parts show average, green above-average and red below-average performance. The last row of the figure illustrates the ground truth disparity maps and the median absolute disparity difference to the ground truth at each individual pixel among all algorithms.

Fig. 12 illustrate the analysis of the 3D models for the "Cotton" image set in Fig.4.f, generated based on the depth maps from the proposed framework, the ground truth and OBER-cross+ANP which is the best algorithm ranked in the benchmark. The purpose of this comparison is to find out how accurate and close the 3D reconstructed data from the estimated depth map is to the ground truth and the most accurate method in the state of the art. Fig.12.b, Fig.12.g and Fig.12.l represent the rasterized color-coded 3D model from the ground truth, proposed method and OBER-cross+ANP, respectively. The color-coded model indicates how close the proposed method is in terms of establishing depth levels to ground truth and OBER-cross+ANP. The transition from red to blue present the area which are closer and far from the camera. By looking at 3D normals in Fig.12.c, Fig.12.h and Fig.12.m one can determine the smoothness of the disparity values estimated by the proposed framework in comparison to the ground truth and OBER-cross+ANP. The Poisson surface reconstruction [50] and the wireframe model which are shown in Fig.12.i and Fig.12.j, outline the capability of the proposed framework in dealing with non-uniform surfaces and following fine structures.

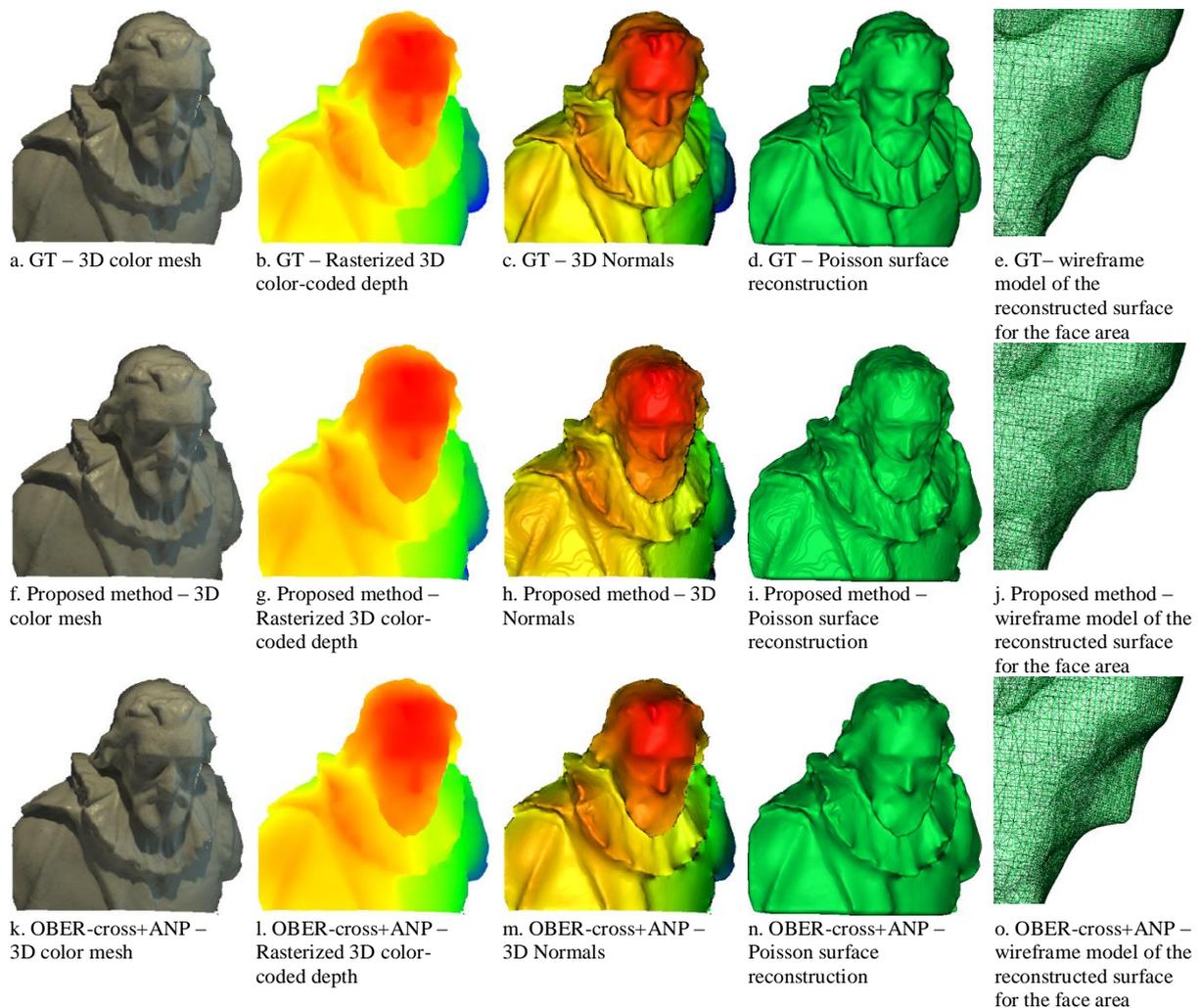

a. GT – 3D color mesh
b. GT – Rasterized 3D color-coded depth
c. GT – 3D Normals
d. GT – Poisson surface reconstruction
e. GT– wireframe model of the reconstructed surface for the face area

f. Proposed method – 3D color mesh
g. Proposed method – Rasterized 3D color-coded depth
h. Proposed method – 3D Normals
i. Proposed method – Poisson surface reconstruction
j. Proposed method – wireframe model of the reconstructed surface for the face area

k. OBER-cross+ANP – 3D color mesh
l. OBER-cross+ANP – Rasterized 3D color-coded depth
m. OBER-cross+ANP – 3D Normals
n. OBER-cross+ANP – Poisson surface reconstruction
o. OBER-cross+ANP – wireframe model of the reconstructed surface for the face area

Figure 12. 3D visualizations of the "Cotton" image set for proposed method, ground truth and OBER-cross+ANP algorithm. First row shows the 3D models based on the ground truth, the second row presents the model generated based on the proposed method and the third row illustrates the 3D models generated based on OBER-cross+ANP algorithm. First column shows the 3D color mesh, second column shows the rasterized 3D color-coded depth, third column shows the 3D normals, fourth column shows the Poisson surface reconstruction and last column shows the wireframe model of the reconstructed surface for the area around the face.

Table 3 and Fig. 13 present the average performance time of the proposed method compared to the studied algorithms in non-logarithmic and logarithmic scale mode, respectively. The computational times for each of these algorithms have been reported by their authors.

A faster performance time and the competitive accuracy of the proposed method make it applicable for deployment in practical embedded systems and Internet-of-Things (IoT) appliances. Using the method proposed in this paper, one could transmit a compressed depth map in an IoT device, rather than the full image stream or analyze the depth map at the edge level and use it to trigger corresponding actions [51]. Table 4 shows how much faster/slower and more/less accurate the proposed method is compared to the other algorithms. For example, the proposed method is ~4.5 times slower than the baseline algorithm EPI2; however, the accuracy of the estimated depth maps has increased ~21% or the proposed method is ~111.8 times faster than SPO-MO but its accuracy decreased ~2.3%.

The estimations in Table 4 are based on the percentage of the pixels with correct disparity values above 0.07 error threshold. Note that, the same metric is initially used to choose these algorithms for comparison purposes.

Table 3. Computational time of the proposed framework and the state of the art in seconds.

| Algorithms | EPI2 | Proposed Method | OBER-cross+ANP | SPO-MO | OBER-cross | OFSY_330/DNR | PS_RF | SPO |
|---|---|---|---|---|---|---|---|---|
| Time (s) | 8.4 | **38.5** | 182.9 | 4304.3 | 96.4 | 200.2 | 1412.6 | 2115.4 |

Table 4. Comparison of the proposed method and the state of the art. Factors of computational time improvement and percentage of disparity accuracy variation.

|  | EPI2 | OBER-cross+ANP | SPO-MO | OBER-cross | OFSY_330/DNR | PS_RF | SPO |
|---|---|---|---|---|---|---|---|
| **Proposed Method vs.** | ~4.5× slower ~21% inc. | ~4.7× faster ~3.5% dec. | ~111.8× faster ~2.3% dec. | ~2.5× faster ~4.95% inc. | ~5.2× faster ~6% inc. | ~36.7× faster ~0.01% dec. | ~54.9× faster ~1.43% inc. |

* inc: Increased accuracy    dec: Decreased accuracy

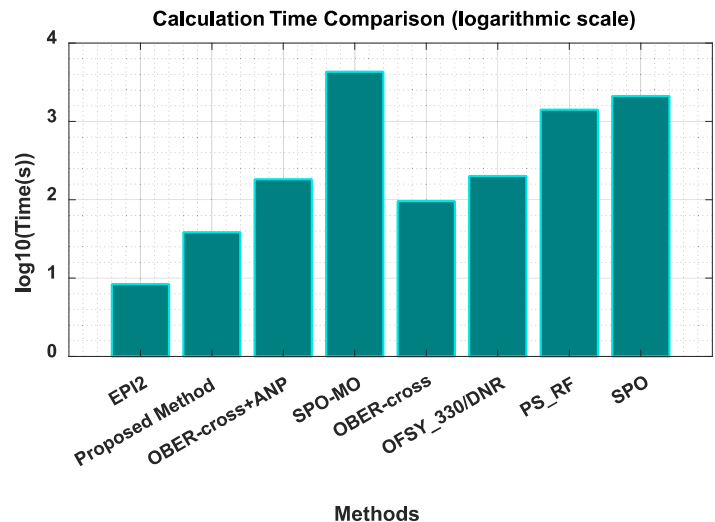

Figure 13. Computational time of the proposed framework and the state of the art in logarithmic scale.

## 5. Conclusion and Discussion

In this paper, a new framework is proposed based on EPI analysis and TV minimization to estimate depth from the multi-camera array. A new cost function is proposed and analyzed based on Fenchel-Rockafellar duality [17]. Our approach consists of three steps. First, the captured sequence of images from the array of cameras is aligned using Epipolar homography alignment. Later, a rough initialization of the depth map is computed using local depth estimation on EPIs. In the third step, this initialization is refined by applying a TV minimization based on Fenchel-Rockafellar duality [17].

We demonstrate the benefits of the proposed framework on a synthetic dataset including Stratified and photorealistic light field image sets. The method has been implemented in Matlab R2017a on a device equipped with Intel i7-5600U @ 2.60GHz CPU and 16 GB RAM.

The evaluation reveals that most algorithms consist of multiple elements and terms which make it difficult to establish one best algorithm that outperforms in all categories. Also, the high computational time of the studied methods makes almost inapplicable for consumer devices. The results demonstrate the competitive performance of the proposed framework among the top state of the art methods in terms of accuracy of depth estimation. Even though the accuracy of the estimated depth maps using proposed framework varies based on each metric, it still remains in the list of high accuracy methods and the fast convergence of the proposed cost function and its fast computational time make it a potential method for consumer electronics applications and devices with the aid of parallel technology and GPUs. The new generation of GPUs contains a high number of programmable parallel cores (up to 4k). This evolution makes this technology an efficient choice for computationally intensive processes in machine vision applications such as depth estimation. We aim to explore the effect of parallelism on depth estimation from the multi-camera array in the future works.